\documentclass{article}

\usepackage{arxiv}
\usepackage{float}
\usepackage{xcolor}
\usepackage{algorithm}
\usepackage{algpseudocode}
\usepackage[utf8]{inputenc} 
\usepackage[T1]{fontenc}    
\usepackage{hyperref}       
\usepackage{url}            
\usepackage{booktabs}       
\usepackage{amsfonts}       
\usepackage{nicefrac}       
\usepackage{microtype}      
\usepackage{lipsum}		
\usepackage{graphicx}
\usepackage{natbib}
\usepackage{doi}
\usepackage{caption}
\title{AIGenC: An AI generalisation model via creativity}

\author{ 
\href{https://orcid.org/0000-0002-9186-3092}{\includegraphics[scale=0.06]{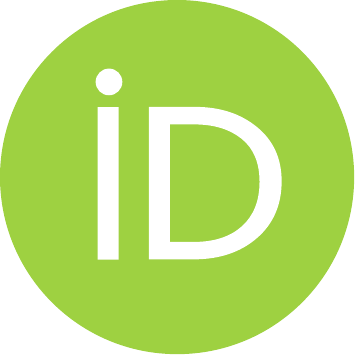}\hspace{1mm}Corina Cătărău-Cotuțiu}\\
	Artificial Intelligence Research Centre (CitAI)\\
	Department of Computer Science\\
	City, University of London \\
	\And
	\href{https://orcid.org/0000-0003-4180-1261}{\includegraphics[scale=0.06]{orcid.pdf}\hspace{1mm}Esther Mondragón}\\
	Artificial Intelligence Research Centre (CitAI)\\
	Department of Computer Science\\
	City, University of London \\
	\texttt{e.mondragon@city.ac.uk} \\
		\And
	\href{https://orcid.org/0000-0002-3306-695X}{\includegraphics[scale=0.06]{orcid.pdf}\hspace{1mm}Eduardo Alonso}\\
	Artificial Intelligence Research Centre (CitAI)\\
	Department of Computer Science\\
	City, University of London \\
}

\renewcommand{\shorttitle}{AIGenC: An AI generalisation model via creativity}

\hypersetup{
pdftitle={AIGenC: An AI generalisation model via creativity},
pdfsubject={q-bio.NC, q-bio.QM},
pdfauthor={Corina Cătărău-Cotuțiu, Esther Mondragón, Eduardo Alonso},
pdfkeywords={AGI, Deep Reinforcement Learning, Generalisation, Learning transfer, Creativity},
}

\begin{document}
\maketitle
\raggedbottom

\begin{abstract}
Inspired by cognitive theories of creativity, this paper introduces a computational model (AIGenC) that lays down the necessary components to enable artificial agents to learn, use and generate transferable representations. Unlike machine representation learning, which relies exclusively on raw sensory data, biological representations incorporate relational and associative information that embeds rich and structured concept spaces. The AIGenC model poses a hierarchical graph architecture with various levels and types of representations procured by different components. The first component, Concept Processing, extracts objects and affordances from sensory input and encodes them into a concept space. The resulting representations are stored in a dual memory system and enriched with goal-directed and temporal information acquired through reinforcement learning, creating a higher-level of abstraction. Two additional components work in parallel to detect and recover relevant concepts and create new ones, respectively, in a process akin to cognitive Reflective Reasoning and Blending. The Reflective Reasoning unit detects and recovers from memory concepts relevant to the task by means of a matching process that calculates a similarity value between the current state and memory graph structures. Once the matching interaction ends, rewards and temporal information are added to the graph, building further abstractions. If the reflective reasoning processing fails to offer a suitable solution, a blending operation comes into place, creating new concepts by combining past information. We discuss the model's capability to yield better out-of-distribution generalisation in artificial agents, thus advancing toward Artificial General Intelligence.

\end{abstract}


\section{Introduction}
AI systems can sometimes struggle to adapt to novelty. More often than not, AI fails to effectively use and exploit past learnt experiences when faced with new challenges. This deficiency can manifest as the inability to apply prior knowledge but also as impaired learning. This shortcoming is not the result of catastrophic forgetting when trained on novel data, but rather it is simply a limitation by design of machine learning algorithms to adapt to changing circumstances.

It can be argued that this limitation is due to AI systems' poor representational capability to acquire meaningful relational concepts that can successfully transfer information across different contexts. For example, in large language models, zero-shot generalisation to previously unseen words is observed, and generalisation to previously trained words suffers in a different context (Dasgupta et al., 2020). 
Such context dependence indicates that AI systems might have to parse each word in a wide variety of embeddings in order to generalise it to new contexts, suggesting that combinatorially large amounts of augmentation will likely be required for a \textit{tabula rasa} unstructured neural network model to learn an entirely systematic representation (Dasgupta et al., 2020).
 
 We hypothesise that Creative Problem Solving (CPS) scenarios could help AI systems learn versatile concepts that can be adapted to novel situations, a step forward to solving the problem of generalisation in AI.

Generalisation is a natural process by which individuals respond in the same way to different but similar information. Human beings and other animals have a formidable capacity to map previous experiences to fresh situations, thus adjusting to their environments efficiently (Momennejad, 2020; Murphy et al., 2008; Wu et al., 2018). Machines do not. Effective adaptation, however, requires, on many occasions, more than passive reuse of information, demanding creative solutions like building a tool with unusual materials. 

As deep networks become more sophisticated and approach human-like performance in certain tasks, serious concerns arise with regards to the nature and lack of structure of the representations they learn and their effect on decision making (Dasgupta et al., 2020). To address these issues, we take inspiration from CPS theory to investigate the requirements and the corresponding modelling components that would allow a naïve agent that learns by trial and error while interacting with an environment, that is, a reinforcement learning (RL) agent, to act efficiently across domains in multiple tasks. This approach diverges from standard deep RL architectures that merely reduce the complexity of the environment by condensing raw input. This strategy does not capture the higher-order cognition needed to enable agents to acquire information that makes sense of the structure of the world.

Stimulus generalisation in the natural domain is driven by the extant common elements present in the input. By virtue of these commonalities, humans detect similarities between different sources of information and are capable of transferring sensory knowledge from one setting to another. Yet, stimulus generalisation relies exclusively on sensory attributes that are often embedded within numerous irrelevant cues, both similar and dissimilar, which may lead to inadequate or dysfunctional use of information. Furthermore, stimulus generalisation is only a fraction of human generalisation capabilities. For instance, associations can also mediate generalisation to dissimilar cues, allowing for the transfer of learning across different sensory dimensions (Eilifsen \& Arntzen, 2021). In addition, the role of commonalities can be extended to different levels of information (Mondragón et al., 2017). Thus, extracting suitable patterns and relationships capable of bearing forward resemblance across situations beyond simple sensory input is fundamentally adaptive as it enables humans to build up compressed information; that is, to progressively reduce otherwise overwhelming stimulation that would saturate their perception, and create hierarchies upon which knowledge can be transferred. Generalisation and abstraction compress our world to manageable proportions, make it familiar and permit transferring knowledge to analogous scenarios (James, 1890).

AI, deep learning, in particular, excels in extracting regularities or patterns from training data within the same-distribution, displaying superhuman performance in specific tasks (Bengio et al., 2021; Lyre, 2020). However, at present, artificial agents are mostly unable to generalise experiences that are significantly different from those already encountered and to transfer previously learnt knowledge to out-of-distribution settings. This begs the question, where does the discrepancy between superhuman performance on one task and spectacular failures on a similar but different job originate?

Deep neural networks sometimes employ shortcut strategies to reach a specific goal (Geirhos et al., 2020). Shortcuts are spurious correlations extracted by networks to solve a task without learning the true structure of the input. They involve data interpolation –meaningless internal ad-hoc adjusting that reduces the error– and are, thus, highly sensitive and dependent on a particular training dataset. Interpolating agents will not, in general, extrapolate, i.e., perform well beyond the training task -unless, of course, supplemented with prior knowledge about the function outside the data range. This obstacle is consonant with difficulties encountered by deep networks when tested beyond domains where large amounts of training that can dissipate the effectiveness of shortcuts are not present. The problem of extrapolating without additional data has been targeted recently by approaches such as continual learning for non-stationary data or transfer learning, neither of which has offered fully satisfactory solutions (Mitra, 2021).

Some successful systems appear to perform as if they had learnt complex and abstract concepts and could potentially transfer learning (e.g., text to image generation (Ramesh et al., 2021), text generation (Bayer et al., 2022; Yang et al., 2021) etc.). This illusion is nurtured by the assumption that human-like performance implies human-like strategies, that is, by the belief that behaving in a manner akin to humans presupposes similar underlying cognitive traits. This fallacy is rooted in the prevalent notion of computationalism and its disregard for the physical substrate of cognition (Searle, 1980). Model performance in artificial neural networks is based on simple correlations, and outputs are usually just a statistical good fit (Floridi \& Chiriatti, 2020). This, as commented earlier, is problematic as the features that minimise error do not necessarily support extrapolation. In addition, the agent's learning experience upon which concepts are to be acquired is limited to single scenarios with features predominantly extracted from a single sensory channel. Consequently, generalisation of performance can only be achieved based on interpolation within the sensory domain of the training set.

Endowing artificial agents with core concepts –abstract representations that capture meaning beyond the training context (Chollet, 2019; Mitchell, 2021), that is, with a comparable capability of building and manipulating abstract information at different hierarchical levels, is essential to achieve human-like performance. In the absence of such concepts, current AI models produce non-optimal responses in unknown environments and rely on large datasets and large numbers of parameters to produce seemingly superhuman results (Reed et al., 2022), as is the case of transformers in the popular GPT-3, whose impressive performance is due to it using 175 billion parameters and 45TB of text data (Rae et al., 2021).

Following Olteţeanu (2020), we define different types of data collectively as concepts –a consolidated unit that enables us to look for similarities and match information parsimoniously. Accordingly, concepts are objects, events and properties – all of which constitute the building blocks of meaning and so-called common sense that allows humans to rapidly understand and interact with novel scenes (Shanahan et al., 2020; Shanahan \& Mitchell, 2022). This is in contrast to current approaches of AI, where concepts are learnt features extracted from raw data, thus lacking the abstraction needed to structure our thinking at a higher cognitive level.

The notion of applying a concept to very similar, matching scenarios is known as near transfer. It involves using the same concept across contexts. However, for a system to display human-like behaviour, it is critical to be able to modify existing concepts and adapt them to new tasks –effectively, to learn a new concept based on previous knowledge in what is called displacement or creative transfer (Haskell, 2000). Therefore, a mechanism for transferring concepts in a way that is more than just shifting them from one situation to another but instead involves the creative generation of new concepts is needed. The more skilled an agent is at transferring a more creative and efficient solution to a new scenario, the closer its behaviour is to that of humans.
Moreover, there are cases where a problem cannot be solved solely based on existing knowledge, and approximating a solution is not viable. In such cases, it is necessary to introduce creative problem-solving and give rise to new concepts. 

CPS moves AI beyond performance benchmarks and introduces a focus on scenarios for which rehearsed solutions are not valid. While benchmarks (e.g., ImageNet, GLUE, Super GLUE) are useful for measuring AI systems' performance, they have limitations. They are designed to work under restricted conditions (e.g., the object must be centred and with a frontal view) and hence cannot capture the full complexity of the world. In these scenarios,   better performance does not convey real progress towards general intelligence and introduces confounders in the use of artificial terminology and its human counterpart, popularising polemic terms such as 'understanding'. Therefore, researchers cannot rely on benchmarks to assess the broader capabilities of AI models (Raji et al., 2021). 

An agent displaying creative thinking should be able to generate novel solutions from existing knowledge (Frith et al., 2021). Although task transfer does not inherently necessitate creativity, task-tailored solutions might be hard to generate when the degree of dissimilarity between two tasks is high. In such cases, creative solutions may be required.

A recurrent misconception is that the implicit relations learned by neural networks and captured in the weight matrices extracted from single sensory information (Doumas et al., 2022), are sufficient to learn concepts with meaningful knowledge (Doumas et al., 2008). However, more comprehensive information needs to be extracted to achieve flexible cross-domain generalisation, such as implicit and explicit representations of relations, affordances and temporal information, which capture the world's content and dynamics. By implementing the characteristics of different types of knowledge, we may bestow agents with something akin to common sense (Shanahan et al., 2020), a capability which would allow them to infer hidden information (e.g., intentions, goals, utility, meaning) that it is not usually detectable by sensory correlations alone. We assume that common sense is embedded in the information that an agent learns by interacting with the environment, which in artificial settings can be simulated using RL (Sutton \& Barto, 2018).

In traditional RL scenarios, agents disregard by design most of the information provided by the environment, learning simple policies (sequences of actions that accumulate rewards) linked to global states. Although, in theory, states in RL can represent any type of information, in practice, RL implementations work primarily on states in which only sensory data is encoded (Badia et al., 2020; Vinyals et al., 2019). In so doing, states are monolithic entities that do not allow for the construction of concepts and their transfer. We argue that to achieve more robust concept representations, an agent cannot rely on raw sensory information only, but rather it must learn by trial and error the functional and contextual information that accompanies it. Thus, we need to exploit these enhanced representations as a first step towards achieving a creative agent that can adapt to new tasks in RL scenarios. Suppose we wish to provide an agent with the ability to adapt knowledge creatively. In that case, we need to expand the RL framework with a mechanism that enlarges the set of actions and makes creative problem-solving viable.

There have been several approaches to enriching RL with CPS capabilities. Recently, Gizzi et al. (2020) developed a formal model that encapsulates states, actions, and policies under a single concept space on which they defined different creativity functions (e.g., combination, transformation). Similarly, Colin et al. (2016) drew a parallel between Wiggins' Creative System Framework (2006) and hierarchical reinforcement learning. The Creative System Framework (Wiggins, 2006) interprets creativity as a concept search within a problem space and a meta-search of problem spaces. Colin et al. (2016) made these two levels of search equivalent to searching the object space in the environment and searching the policies in the RL space. Despite the theoretical relevance of these approaches, they are high-level formal proposals which do not offer specifications on how to integrate CPS and RL. Our work aims at formulating a computational framework for machine learning implementations that incorporate CPS in a deep RL environment.

In this paper, we introduce a theoretical computational model inspired by cognitive theories of CPS that, we argue, would enable artificial agents to learn heterogeneous, generalisable concepts and transfer relational information. We postulate an adaptable concept space that encodes objects, affordances and relational information from environmental features, allows for a targeted adaptation of its contents and hierarchically builds up an abstract temporal layer that captures the dynamics of the environment. The main characteristic that sets aside our framework is positing different dimensions of knowledge (sensory, dynamic, temporal and relational) for matching information at different levels of abstraction. Moreover, we incorporate explicit relations between dimensions to enable the creation of new concepts through which, by process of comparison, could potentially transfer. In other words, we conceptualise a template for a biologically plausible data representation to facilitate efficient cross-domain learning.

The rest of the paper is structured as follows: section 2 covers the principles that sustain the theoretical aspects behind our framework, as inspired by cognitive science. We will elaborate on the characteristics that a functionally creative agent should display, on the concept space and the importance of dynamic information for transfer. Section 3 establishes the link between testing concept transfer and creative problem-solving. Section 4 introduces the main structure of the framework, namely Concept Processing, Reflective Reasoning and Blending embedded in a deep reinforcement learning model. Finally, we conclude with a discussion regarding generalisation in the context of Artificial General Intelligence (AGI).

\section{Functional creativity, concept space and affordances}
The problem of creativity has been approached by different disciplines. Our purpose is to draw inspiration from cognitive models of creativity to implement a more robust and transferable way to learn and solve problems in artificial agents. We argue that a creative problem-solving approach will lead to better AI generalisation.

Humans begin exploring the world with a limited set of concepts, but as the need arises, we learn new concepts, progressively increasing their level of abstraction and forming categories that filter and reduce their dimensionality (Sloutsky, 2010). The ability to build new concepts is associated with functional creativity.

Concepts in the philosophical and psychological traditions are essentially definitions that include enough features to characterise them as belonging to one category or another. That is, concepts are described as entities with the necessary and sufficient conditions for assigning membership of concept X to category Y (Coraci, 2022).
A common assumption in classic creativity theories (Fauconnier, 1998; Gärdenfors, 2004; Mednick, 1962; Olteteanu, 2020; Wallas, 1926) is the requirement of a concept space, which serves as a bridge between sub-symbolic and symbolic representations. In line with these theories, the framework for creative problem-solving that we are proposing in the following section is built upon the notion of concept space and affordances.

Gärdenfors (2004) introduced a conceptual space framework for modelling categories, properties and concepts using a geometrical and topological approach to concept representation. According to the conceptual space framework, concepts are sets of properties belonging to multiple dimensions. Learning modifies these quality dimensions, which thus evolve over time and are ascribed to different domains.

\begin{figure}[H]
    \centering
    \includegraphics[height=10cm]{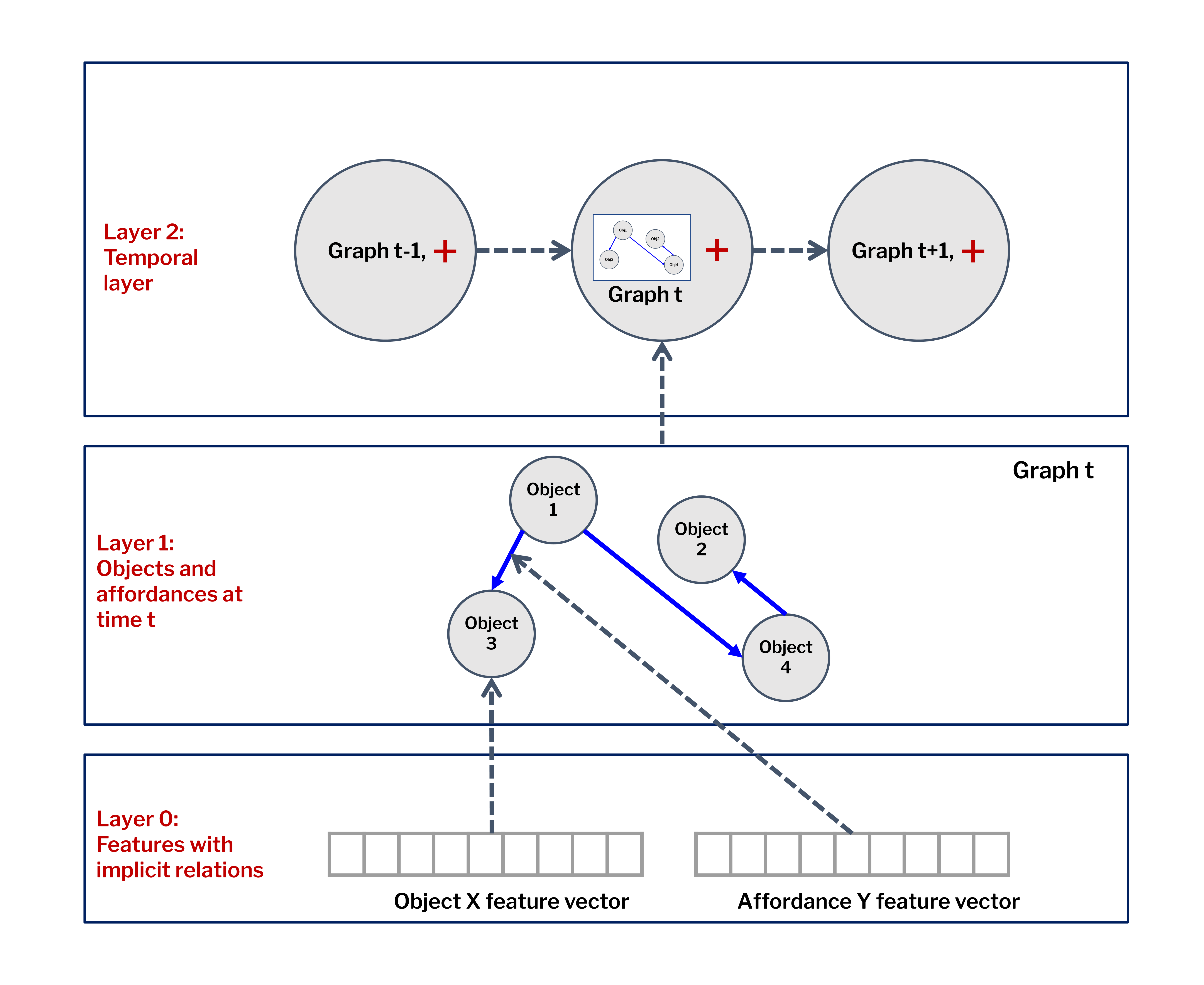}
    \caption{Hierarchical concept space: object concepts are represented by feature vectors at the bottom level (Layer 0). At the middle level (Layer 1), nodes and edges represent object concepts and affordances, respectively. At the top level (Layer 2), the graphs from the previous level become nodes along with a reward, given to the agent in the RL setting, (red +) in a graph whose edges mark sequential information (t).}
    \label{fig:cs_space_hier}
\end{figure}

One of the main hurdles with Gärdenfors' view when adapting it to an artificial setup relates to the origin of the features or dimensions. In artificial agents, a finite initial set of features can be provided manually by a knowledge engineer (Russell \& Norvig, 2002). However, bottom-up concept building becomes unscalable as the complexity of the problem and the number of features increase. Deep learning tools could allow an agent to extract the features pertinent to a given task and scenario in an unsupervised manner and to build concepts based on those. This will remove the intermediator and provide the agent with a mechanism to filter an unlimited set of possible features. Yet, as the world becomes more complex, the harder it is to define concepts in terms of features alone. Representing the concept space hierarchically, such that the combination of low-level concepts enables the definition of more complex ones, is thus essential.
We define concepts as entities formed of several features, and we conceptualise higher-level concepts as combinations of low-level concepts instead of traditional categories. The idea is to employ a single unit (the concept) as a structure that builds up hierarchically and for which the same algorithm can be applied at different levels of abstraction. The hierarchy (Fig.\ref{fig:cs_space_hier}) builds as follows: static concept features as those captured in standard deep RL frames are at the bottom (Layer 0). Above it (Layer 1), base object concepts are represented as nodes of a graph whose edges establish their relations as affordances. Next, at the top (Layer 2), we have a higher-level graph representation whose nodes reproduce the previous graph structure, and the edges capture their temporal succession.

In building such a hierarchy, the first step is finding the properties that would form the base concepts. As the human operator (knowledge engineer) is replaced by an autonomous agent, the system should be capable of sensing the environment to capture the features that would form these primitive concepts, without the aid of an external agent. Such base concepts would be compressed data in a latent space. Data compression techniques are used to generate encodings that represent meaningful information in a reduced form. This is important because storing complete images or auditory patterns as numbers is computationally expensive. Instead, these encodings exist in what is called the latent space representation, which allegedly contains all the relevant information needed to represent the original data points in a simplified way.

The quality of the latent space representation affects how well a model performs and generalises. In a reinforcement learning scenario, for example, identifying which environment features are valuable for encoding the latent state can improve the accuracy of predicting future rewards. This is an active area of research, with recent studies exploring ways to optimise the encoding process for better performance (Radulescu et al., 2021).

For the purpose of this model, the dimensions of the latent space form base concepts. The latent encodings represent low-level sensory features that capture their implicit relations, present in the data itself and learned by a neural network. Implicit relations are also coded as weight matrices that act as relational units between the different features. A clear advantage of using implicit relations is that they capture functional relationships directly impacting the model's behaviour (Doumas et al., 2022). However, being implicit, the latent encodings cannot be changed without changing the representation itself. Therefore, although we continue using implicit relations for sensory features, we posit the need of encoding explicit relations for inter-concept interactions.

First, we need the agent to learn in its interaction with the task and environment the necessary information critical to the task and form base concepts.  

Unsupervised learning thus channels the agent's discovery of hidden patterns without the need for human intervention, filtering out external biases. It serves to discover inherent structures, as the goal is not to predict a specific outcome but to obtain insights from the data. Hence, unsupervised learning of basic concepts should be the first component of our framework. 

Of particular interest are unsupervised representation learning techniques like autoencoders (AE) (Masci et al., 2011; Rumelhart et al., 1986) and variational autoencoders (VAE) (Kingma \& Welling, 2014; Rezende et al., 2014) because
they learn low-dimensional representations of high-dimensional distributions by encoding and decoding the input data, holding a promise of inducing features automatically from raw perceptual data.

Up to this point, we have focussed on one dimension provided by sensory information. However, as we have emphasised earlier, sensory information alone does not suffice to achieve efficient learning transfer. Having different dimensionalities of knowledge is key in coding useful concept representations. When presented with a problem-solving task involving objects, children usually would first interact with them to find out what they are and what can be done (Xie et al., 2021). How an object can be used is known as its affordance. Gibson (1977) introduced the term \textit{affordance} to describe the idea that certain states enable an agent to perform certain actions in a given context and that they exist as a combination of properties concerning a specific agent. Formally, affordances are defined as relations in the agent-environment system rather than as properties (Chemero et al., 2003).

Affordances and creativity are intertwined notions, especially in the case of functional creativity, where the initial concept space becomes insufficient for solving a problem. Sensory similar concepts are not bound to behave the same way. A beach ball and a  bowling ball may look similar, but their weights will render them functionally divergent to, e.g., balance a lever. An agent must be able to interact with objects and, by trial and error, assess their possible uses to learn their affordances.

We, therefore, propose enriching the formation of a hierarchical concept space by including affordances that would relate to the encoded physical features of objects. Incorporating affordances will enable an agent to form various complex representations in a given functional, spatial and temporal context, hence acquiring knowledge about object manipulation.

Implicit relations capture the characteristics of actions as base concepts and along with the relation that the action establishes between different concepts, it defines an affordance, thus encapsulating the dynamics of the world.

The resulting structure can be represented as a graph, whose nodes are concepts and edges are affordances. 
Whereas sensory features can be represented as a whole, their temporal and implicit relational dynamics must allow explicit modification and hence be represented distinctively as separate concepts. Similarly to the sensory features, the action component of affordances can be learned in an unsupervised way. The affordance concept would then be defined as a vector representation of basic concepts and a relation. Since our purpose is to design an agent capable of transferring concepts across different settings, the agent must be able to alter how objects interact without altering the concepts themselves. Hence explicit relational information that incorporates the meaning of the relationship and is bound to other concepts must be represented independently.

The hierarchical structure we envision aligns with Olteteanu's (2020) research, but instead of resorting to predefined knowledge bases, we advance a creative problem-solving algorithm that manipulates concepts in the latent space using deep learning methods.

\section{Assessing concept transfer in creative problem solving}

Theoretical approaches in cognitive science often distinguish between two processes of creative thought. One process, known as divergent thinking, is characterised by searching for a distinctive novel solution to a problem that has multiple correct answers (open-ended). Divergent thinking is usually assessed by exploring the space of possible solutions (Guilford, 1967). On the other hand, creative problem solving (CPS) is a process directed at solving ill-defined problems that do not have known operators and solutions, and that often require task reframing. We argue that CPS is a strategy that would allow an artificial agent to come up with a novel use for a known object, a means to achieve generalisation and knowledge transfer (see also, Jacobs \& Dominowski, 1981). Solving a problem may require operating a particular tool not present in the environment. An agent facing such a scenario may need to search for a substitute or devise new objects by putting together others (or their parts) available to the system. 

Creative use of learnt affordances that encapsulate the dynamics of the world can solve these types of problems (Oleteanu, 2020). In our model, affordances are an integral part of the concept space's hierarchical structure, which holds and compartmentalises concepts at different levels of abstraction that can be manipulated individually. Such a structure would facilitate the transfer of affordances (concepts) independently of the object from which they were extracted. 

A hierarchical concept space from which creative outcomes can be produced needs to be built from a sufficiently rich medium such that enough concepts can be gathered to support generalisation. RL environments provide us with an ample sample of concepts that the agent explores in frames of increasing complexity, forcing it to learn the dynamics of the world independently of specific sensory features to complete a given task (e.g., Jain et al.'s (2020), Beyret et al.'s (2019)).

 \section{A framework for concept transfer and functional creativity}
 
It is important to note that this theoretical framework seeks to lay the fundamental structures and interplay necessary for good learning transfer in RL systems by looking at cognitive models for insight, not to present a particular implementation. This section describes these units and interactions in detail.

\begin{figure}[htp!]
    \centering
    \includegraphics[width=\textwidth]{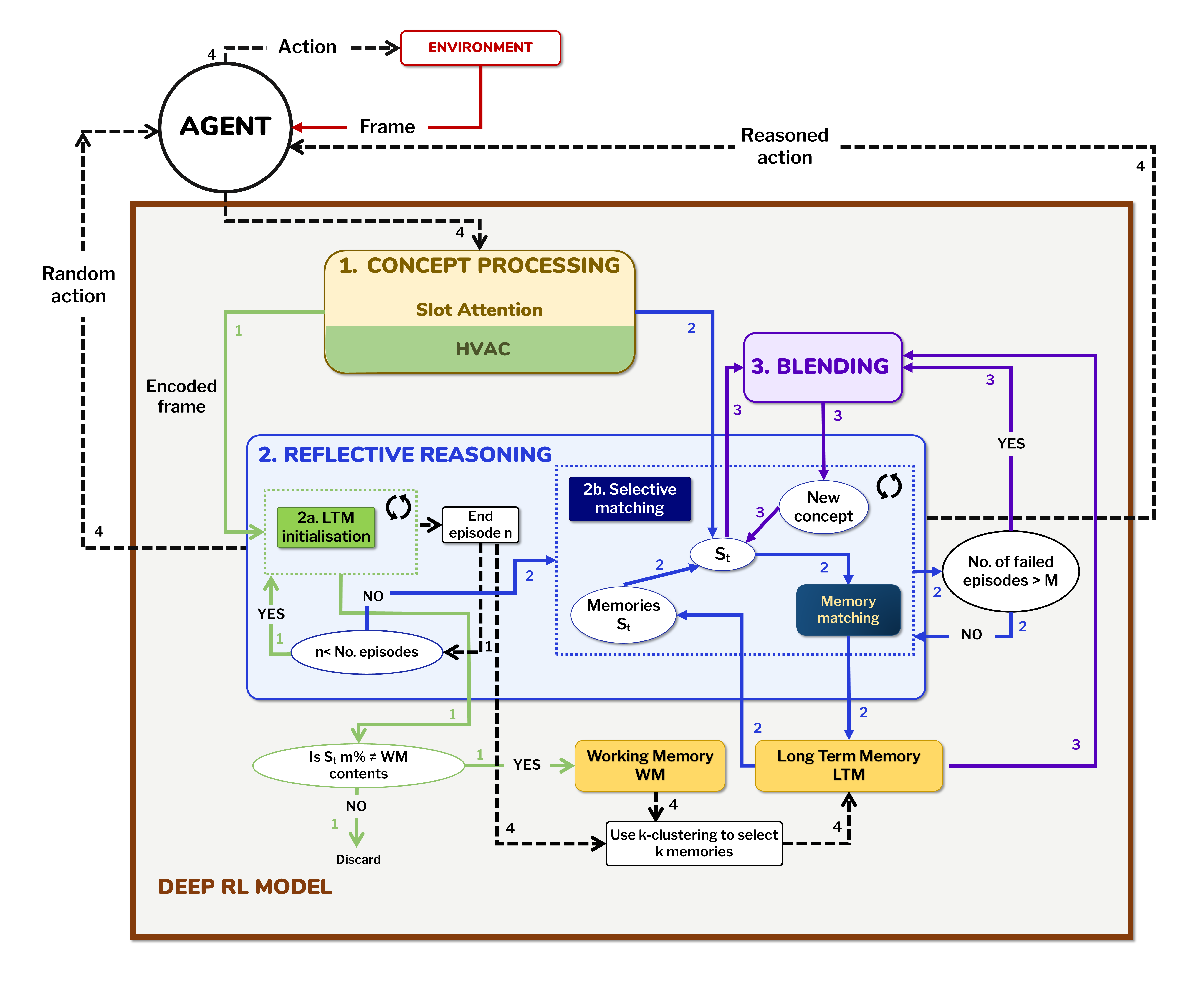}
    \caption{The three components of the model - Concept Processing (cream/olive), Reflective Reasoning (blue), Blending (purple) - and the algorithmic flow. The agent receives input in the form of frames from the environment. Frames are encoded in the Concept Processing component as vector representations by two unsupervised models. In the Reflective Reasoning component, LTM is initialised by random exploration (2a). The process of populating Working Memory and LTM is indicated by open tip arrows, a solid green (1) and a dashed black line (4), respectively. Following LTM initialisation, a process called Selective Matching (2b) is activated together with Deep Reinforcement Learning training. Selective Matching returns from LTM the concepts most similar to the current state. The retrieved concepts are then incorporated into the current state before they are inputted into the Deep Reinforcement Learning model. If the agent is unsuccessful at solving the task for several episodes, a Blending process is triggered, by which novel concepts are formed and then added to the current state (3).}
    \label{fig:flow}
\end{figure}

The model posits a three-component deep learning structure and the respective operations leading to functional creativity. It includes Concept Processing, Reflective Reasoning, and Blending. Of these three high-level components, the first two comprise two sub-components each: Concept Processing involves object discovery and affordance learning, and Reflective Reasoning consists of Long Term Memory (LTM) initialisation and selective matching (Fig.\ref{fig:flow}).

The framework runs on top of a basal goal-directed deep RL training that relies on a hierarchical concept space. The latter's structure exploits both implicit and explicit relational information to support flexible knowledge transfer. To save resources and acknowledge the influence of phylogenetic and ontogenetic evolution, implementations may need to pretrain the representational architecture.

The first component, Concept Processing, learns different representations extracted at each time-step. Such representations, constitute a RL state. States are then added to a Working Memory (WM) unit based on their novelty and used together with Long Term Memory (LTM) by the Reflective Reasoning component to select the next action. The term \textit{reflective reasoning} - borrowed from Nickerson (1987) and Fletcher and Carruthers (2012), defines the main functionality of the second component, namely, a process of comparison between stored knowledge and current input intended to adaptively select and combine valuable information for the task at hand. Finally, the agent creates new concepts using the Blending component when available concepts do not overcome a standstill in solving a task.

\subsection{Deep reinforcement learning}
At its core, any reinforcement learning task is defined by three features - states, actions and rewards. RL agents learn by receiving direct feedback. Learning a task (goal) is accomplished by accumulating rewards distributed by the designer to guide the interactions of the agent with the environment (Lyre, 2020), unlike other machine learning techniques that rely on data (be it labelled or not). States are representations of the world - they inform the agent 
of which situation it currently is in. An RL agent learns to optimise a policy to reach a goal state, by maximising the cumulative reward; in other words, it learns a strategy guided by the distribution of rewards to pursue its goals. The agent learns such policy by trial and error, exploring the environment and exploiting the knowledge it acquires about the rewards of state-action pairs.

Deep reinforcement learning is a subfield of machine learning that combines reinforcement learning and deep learning (DL). A DL model is a neural network consisting of multiple layers of interconnected nodes. Deep RL agents process the environment by passing frames to DL models as input and choosing an action based on those encodings. Unlike in standard RL, where agents learn policies tied to global states, deep RL agents learn a plethora of environment features. Some deep RL models train the neural networks responsible for learning environment representations together with the RL task in a single loss function (e.g., Baker et al., 2019; Jaderberg et al., 2016; Raileanu et al., 2021). However, other works (Stooke et al., 2021; Yarats et al., 2021) and our proposal isolate the representation learning task from goal-directed training. 
Representations are therefore extracted using pretrained models, which are only updated inside the RL training loop as opposed to trained from scratch. This allows for faster training, removing the need to optimise for two tasks simultaneously. Learning representations within the RL training loop is difficult as the only source for supervision is the reward, which is noisy, sparse and often delayed (Zhang et al., 2022).

The way the representations are formed, pretrained or not, is also essential as simple encoding of full frames disables the agent from distinguishing between core features (i.e., features that serve as discriminative cues that control learning) and features that are irrelevant to the task at hand  (e.g., image quality, resolution, and colours), and forgoes most of the temporal data required for complex multi-level representations (Hayman \& Huebner, 2019).
Our model broadens the scope of deep RL with a setting aimed at distinguishing between irrelevant environment features and concepts that aid in achieving a goal. We outline a mechanism to filter and extract granular and varied information (sensory, functional and relational) to form core concepts transferable across domains. The framework decomposes the sequence of frames into multiple layers of the concept space, capturing objects, affordances, and temporality. All these levels of information are stored in a memory system. In so doing, the agent no longer relies on current static frame data to make its decisions.

A deep RL environment is an ideal experimental setting for functional creativity. It supplies agents with state frames of increasing complexity and, potentially, large samples of concepts that they can learn through both observation and interaction.

To summarise, we posit that a framework capable of functional creativity and transfer first decomposes frames from an RL environment into varied types of information (concepts), which are first stored in a Working Memory unit. The acquired knowledge is then used during the RL training to help the agent navigate and solve a task. At the end of an episode, the \textit{k} most representative concepts are selected and stored in the Long Term Memory unit. The details of how the different components interact are explained in depth in the following sections.

\subsection{Concept processing component}

A concept processing unit is needed to encode and add the input to the concept space. To separate sensory from dynamic information, we postulate two corresponding subunits, both unsupervised and pretrained.

Unsupervised models are used to guarantee that the concept space is internal to the agent, in the sense that the discovered concepts are not predefined by externally given classes, and to endow the agent with the ability to extract representations of objects/concepts regardless of whether it has experienced them before or not. 

The first sub-component aims at finding and representing objects as latent vectors through unsupervised object discovery, while the second at encoding action vectors.
The interaction of the two subcomponents and the environment is described based on a modification of Şahin and collaborators' formalism (Şahin et al., 2007), so that affordances can be defined as an acquired relation between an \textit{(effect, reward)} pair and a \textit{(concept-object, action)} tuple. In this manner, when an agent applies an action to the object, an effect and associated reward pair is generated.

To ensure that concepts are detachable and transferable the concepts in the object discovery component must be extracted in an unsupervised way using object discovery methods, such as autoencoders(AEs), and they need to be filtered. Limiting the number of objects processed is needed to reduce the potential vast information irrelevant to the task, avoiding computational explosion. To filter out the encodings that best characterise the input, we need to select \textit{K} most valuable vectors out of \textit{N} given, thus reducing the search space of possible concepts to only the most salient ones. Filtering based on saliency can be done using a slot attention module (Locatello et al., 2020) on top of an autoencoder architecture. Slot-attention uses an iterative attention mechanism to map the autoencoder latent representations to \textit{K} slots. 

\begin{algorithm}
	\caption{Concept processing integrated within an RL setup} 
	\begin{algorithmic}[1]

 \State Intialising LTM $ltm$ = <keys[], {}>
\State Let $object-discovery$(frames from the environment) $\rightarrow Slots^{(KxM)}$ be a pre-trained object discovery model, where K is the number of slots and M the size of the encoding vector
\State Let $action-encoding$(action representations) $\rightarrow Actions^{KxP}$ be a pre-trained action encoding model, that takes as input sequences of frames in which the agent interacts with an object and returns K vector encodings of size P of those interactions
\State Let $A$ be the set of objects interacting with the agent in the environment
\State Let $create-state-graph$($s_{t}$,$ltm$) be a function that returns the current state graph

\For {$iteration=1,2,\ldots,N$}
 \State Intialising WM $working-mem$ = <keys[], {}>
    \State Let $A_{i}$ be a subset from a set of given random actions that the agent can choose from to solve the given task
    
        \While {$episode$ not done}
            \State Let $s_{t}$ be the current visible state of the environment
            \State objects $\leftarrow object-discovery(s_{t},A_{i})$
            \State action-encodings $\leftarrow action-discovery(A_{i})$
            \State $G_{t} \leftarrow$ create-state-graph(objects,action-encodings, $s_{t}$)
            \If {memory initialisation phase} \State $a_{i,t} \leftarrow$ $random(actions)$
            \Else {} \State $a_{i,t} \leftarrow $ Run policy-network $\pi (G_{t})$  
            \EndIf
            
            \State $s_{t+1}$,reward $\leftarrow$ Take action $a_{i}$
            \State Compute effect $f(a_{i}, s_{t}, s_{t+1}) \rightarrow$ encoding($s_{t+1}$) - encoding($s_{t}$) 
		\State Update $G_{t}$ with ( $s_{t}$, $s_{t+1}$, effect)
            \State Update WM
            
			\EndWhile
			\State Optimise policy network
	
		\EndFor
	\end{algorithmic} 
\end{algorithm}

While for the first sub-component deep learning has achieved remarkable results, there is still not a satisfactory solution for unsupervised affordance learning. For the concept space to be internal to the agent, the affordance learning algorithm has to use feature learning and be unsupervised. The results presented in a recent survey on affordances (Hassanin et al., 2021) show that there are few unsupervised feature learning methods in the literature. A formal model for affordance learning can be defined by incorporating affordances in a relational concept space using a Hierarchical Variational Autoencoder (HVAC) architecture (Edwards \& Storkey, 2016; Jain et al., 2020).

An HVAC, which is a variant of VAEs, is designed to encode sequential data and group observations belonging to one sequence of frames in the latent space. While HVAC captures the evolution of the interaction, it has a limitation in that the encodings of the dynamics are not disentangled from the objects interacting. While this limitation does not require addressing for our agent, as affordances are defined as a combination of agent and action properties, it is a factor that may need to be considered in other settings. Therefore we will not refer to the output of the HVAC model as an affordance but as an action representation. The action representation, together with the effect, will form the affordance (See Algorithm 1).

\subsubsection{Concept space}

The concept space is a hierarchical data structure formed from the representations extracted at the Concept Processing unit. This structured data is necessary for matching and selection of concepts at the Reflective Reasoning component. 

Following Olteteanu's suggestion (Olteteanu, 2020), knowledge is structured as concepts in a three-level encoding: feature space, concept level and a graph representation of an RL state-time configuration defining the problem template.

The concept space is characterised as a graph because the structure should be able to store and relate any kind of information without a predefined data design and satisfy some theoretical prerequisites such as being hierarchical and adaptive (Fauconnier, 1998; Gärdenfors, 2004; Mednick, 1962; Wallas, 1926). Graphs are non-linear data structures consisting of nodes and edges that can be easily manipulated by adding or removing edges without altering the whole graph. They are the most promising means for representing a dynamic concept space able to capture the relations (i.e., affordances as edges) between different elements (i.e., concepts as nodes) and alter those relations flexibly. Graphs can also represent different levels of hierarchical abstraction, such that a graph can be labelled and become a node in another graph. 

Graph representations allow for flexibility, both vertical (i.e., changes in levels of abstractions in the hierarchy) and horizontal (i.e., expansions within a level by adding or removing nodes). In addition,  dynamical graphs (Danks \& Plis, 2019) also permit the kind of flexibility imposed by contextual and time-dependent functional classes in creative problem-solving. Through time and learning, the structure of dynamic graphs is constantly changing. Using such a concept space would require storing  this dynamic graph representation in a memory system. Traditional memory based networks (i.e., Neural Turing Machines (NTMs) (Graves et al., 2014)) operate on matrices. However, the changing nature of dynamic graphs imposes a formal representation based on lists. Adjacency lists can therefore be used to represent a graph as an array of linked lists. Such an array allows economically storing and modifying of complex graph structures in which every node has a list of nodes to which it is related. These lists can be of different sizes, which allows for partially connected graphs.

\subsubsection{Memory system}
Our framework integrates two memories, a Working Memory (WM) and a Long Term Memory (LTM) unit. These two memories must interact throughout the lifetime of an RL agent.

\begin{figure}[H]
    \centering
    \includegraphics[height=10cm]{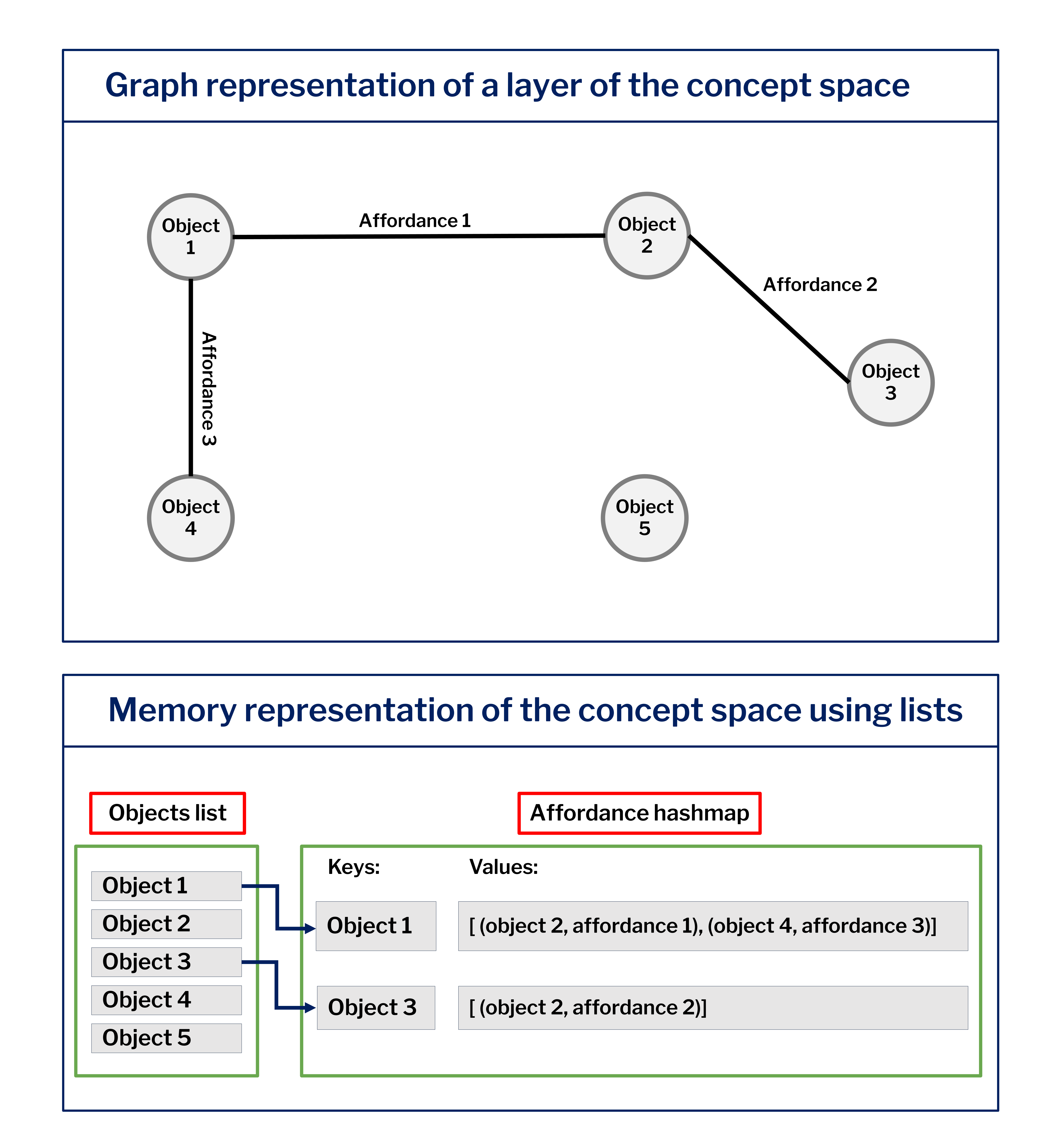}
    \caption{The top panel shows the concept space represented as a mathematical structure, a graph. The bottom panel displays how graphs are coded in memory by using two lists: an object list and a hashmap of affordances. The keys from the hashmap represent the objects (e.g., Object 1), and the values are tuples indicating the connected node (e.g., object 2) and the edge between two nodes, which is an affordance (e.g., affordance 1).}
    \label{fig:cs_lists}
\end{figure}

As advanced in the previous subsection, the memories' content, the concept space, is assembled in the form of two lists (Fig.\ref{fig:cs_lists}). First, an object list records persistently the object representations extracted in the Concept Processing component.
 
Second, a hashmap list, that is, a data structure that maps keys to values, where each key is given by an object of the object list mapped to a value corresponding to an affordance.

Each of the two memories has a particular purpose. First, WM stores current information at each episode in an RL setup. New state information is added to WM or discarded if it already exists (Algorithm 2). Once the agent reaches the terminal state (when the goal is achieved or after a time limit is exceeded), the episode ends, and WM is cleared, at which point the most representative states (the centroid of a data cluster) and their associated concepts are selected from the WM unit using k-clustering and stored in LTM. As opposed to WM, the content of the LTM must persist throughout episodes, to permit transfer of learning during the lifetime of an RL agent (Loynd et al., 2020; Mezghani et al., 2014) (Algorithm 2).
\begin{algorithm}
	\caption{Flow of information through the main algorithm} 
	\begin{algorithmic}[1]
 \State Intialising LTM $ltm$ = $<$keys[], {}$>$
\State Let $memory-matching$($mem$,$G_{t}$) $\rightarrow$ True/False be a function that checks if current state representation already exists in memory
\For {$iteration=1,2,\ldots,N$}
 \State Intialising WM $working-mem$ = <keys[], {}>
        \While {$episode$ not done}
            \State Let $G_{t}, a_{i,t},$ effect$_{t}$ be the variables interacting with the memories, from Algorithm 1. 
             \If {$memory-matching$(working-mem,$G_{t}$)} add to $working-mem$
            \Else {} discard 
            \EndIf
			\EndWhile
			\State Run k-clustering on $working-mem$
                \State centroids-graphs $\leftarrow$ extract-centroids($working-mem-clusters$)
                \State Run $memory-matching$($ltm$,centroids-graphs) and Update $ltm$ with centroids-graphs
	
		\EndFor
	\end{algorithmic} 
\end{algorithm}

\begin{figure}[htp!]
    \centering
    \includegraphics[width=\textwidth]{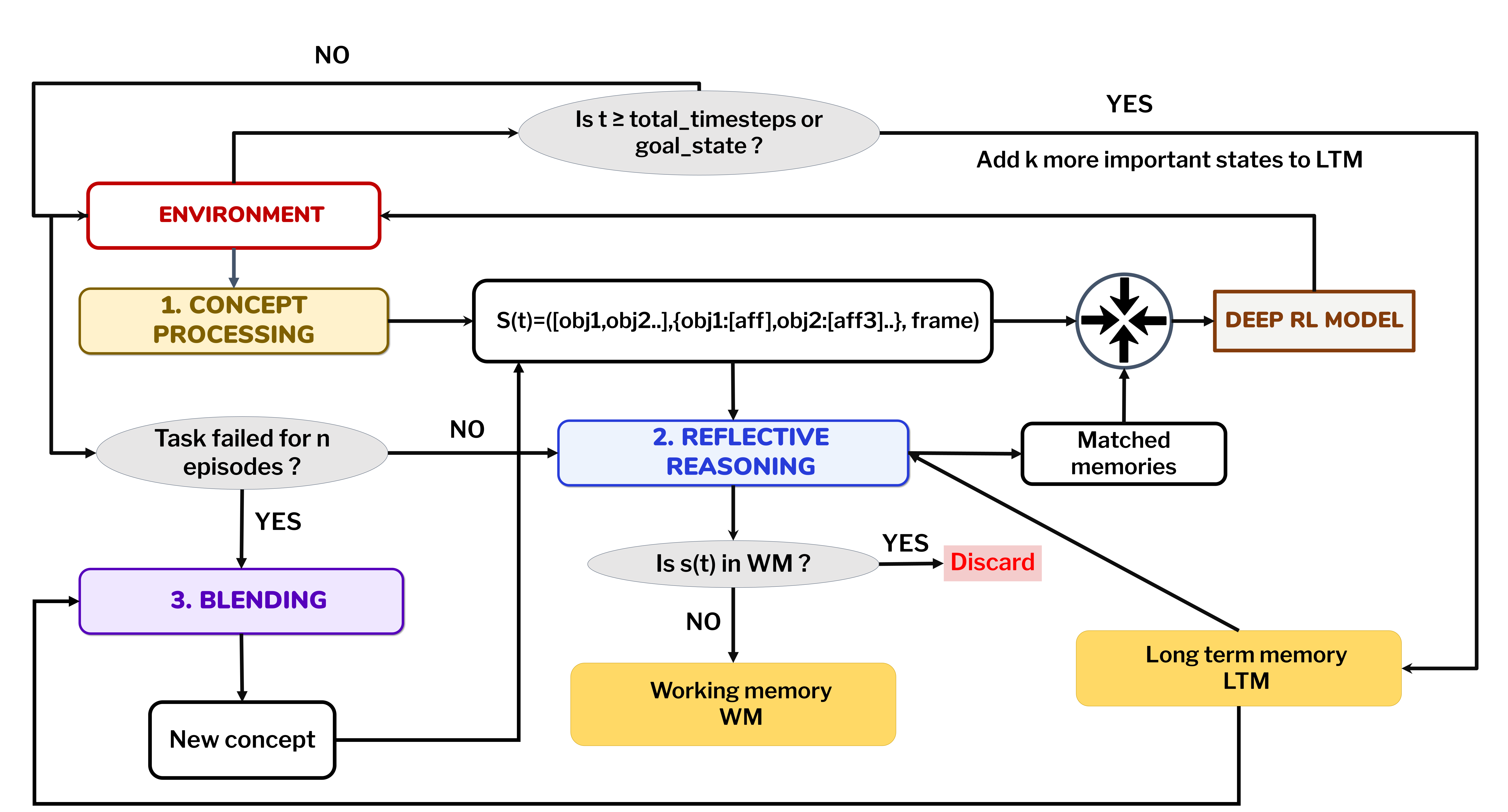}
    \caption{A snapshot of the flow of information through the system. The current state, encoded by the Concept Processing component, is first compared to the concepts stored in Long Term Memory at the Reflective Reasoning component. The concepts that match are used to expand the current state graph. The enhanced current state is then passed to the Deep RL model and applied to the environment. A condition set on the grey top ellipse determines whether the loop is repeated; if the condition is not met, the k most important states from Working Memory are added to Long Term Memory. At the end of a loop, the current state is added to Working Memory if it is different from its current content (bottom ellipse). If several sequential episodes end in failure, Blending is used to enhance the current state with new concepts created by retrieving and combining concepts from LTM (left ellipse).}
    \label{fig:flow_mem}
\end{figure}

Once the concept space is formed, an agent can interact with the stored concepts throughout its creative learning process in both of the remaining components of the architecture, Reflective Reasoning and Blending.

\subsection{Reflective reasoning component}

\begin{figure}[H]
    \centering
    \includegraphics[width=\textwidth]{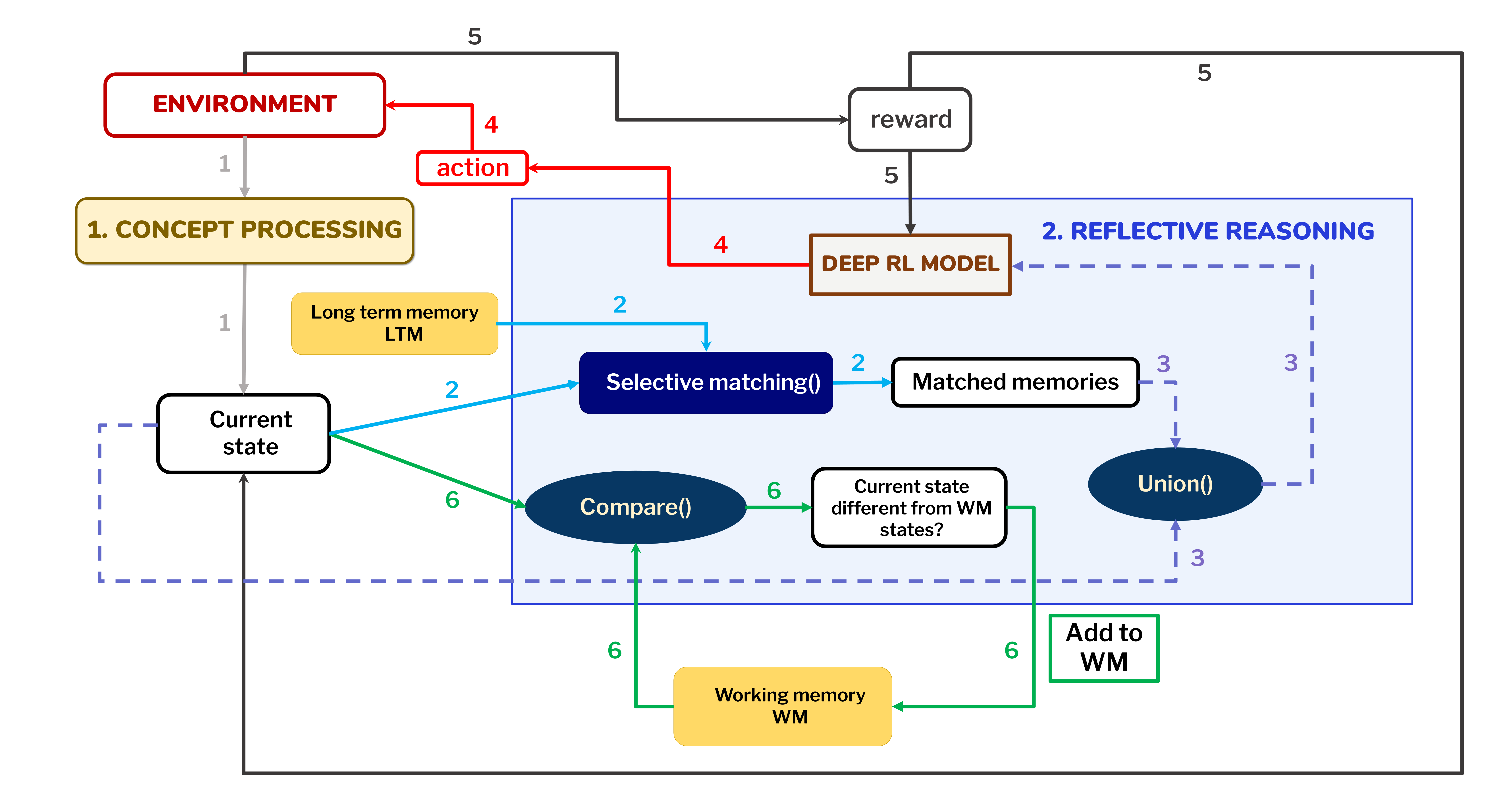}
    \caption{The processing of information within the Reflective Reasoning component is shown through numbered coloured arrows. First, the input is encoded and represented as a graph (1). The processed current state is passed to the memory matching function (2), which outputs the best matches between the current state and Long Term Memory concepts. The concepts retrieved represent the union of the best matches and supplement the current state (3), which is input into a deep RL model. The deep RL model outputs an action which is applied to the environment (4). The deep RL model is updated with a reward and the current state becomes associated with it (5). The updated state is added to Working Memory if it is significantly different (6). The whole process runs in a loop until a final state is reached or a given number of time-steps have passed.}
    \label{fig:comp2}
\end{figure}

Reflective Reasoning, the second component of our framework (Fig.\ref{fig:comp2}), proposes a procedure to choose which concepts from WM are to be permanently stored in LTM and describes the matching operation between the current state and LTM information (see Algorithm 2). We call matching the operation of selecting the concepts useful to fulfil the specific task at any given time-point.

Processing in the Reflective Reasoning component is executed sequentially. First, LTM is initialised; then, selective matching takes place. These two stages and their operation are shown in (Fig.\ref{fig:flow}). The agent initially randomly explores the environment; at each state, it learns new concepts that are loaded into working memory, before permanently storing the most representative ones in the Long Term Memory unit (in this exploration phase the agent uses a random action selection policy, Algorithm 1). Concepts that reach Long Term Memory are then selectively retrieved to enhance the agent's ability to choose an action that can lead to a state closer to the goal in the Selective Matching unit (Fig.\ref{fig:flow}, 2b).

\begin{algorithm}
	\caption{Enhancing algorithm with selective matching} 
	\begin{algorithmic}[1]

\State Let selective-matching($G_{t},ltm, s_{t}$) be a network that returns an updated current state graph with information useful for the task
\For {$iteration=1,2,\ldots,N$}
 \State Intialising WM $working-mem$ = <keys[], {}>
        \While {$episode$ not done}
           \State ...
           \State $matched-graphs$ $\leftarrow$ selective-matching($G_{t},ltm, s_{t}$)
           \State $enhanced-graph \leftarrow supplement(G_{t}, matched-graphs)$  
\State Where supplement is a function that enhances a given graph with $G_{1}$ with nodes and edges present in $G_{2}$ but missing in $G_{1}$
             \State $a_{i,t} \leftarrow $ Run policy-network $\pi (enhanced-graph)$ 
		\State ...
           
			\EndWhile
			\State Optimise policy network
	
		\EndFor
	\end{algorithmic} 
\end{algorithm}

The memory configuration equips the architecture with the means to collect and present structured representations of concepts and to store them in two systems that interact, enabling a process of matching novel states to past information (Doumas et al., 2022). 

Matching, however, does not convey retrieval sameness. Creative agents cannot simply pick exact stored concepts and apply them to a new task. Instead, we need them to adapt existing information, connecting relational concepts (affordances) to novel objects. In this context, CPS demands adjusting previous concepts to the current unfamiliar state (Shanahan \& Mitchell, 2022).
The hierarchical structure of the concept space makes this adaptation possible because it allows independent access to each level of abstraction. The low-level information that comes directly from the Concept Processing component is stored alongside higher-level representations that abstract spatial, temporal and context information from the frame.

Matching nonetheless implies a process of comparison to assess similarity by measuring the distance between elements involved. Since the concept space and the states are represented as graphs, comparisons must be made on graphs. Analysing and comparing graph data is challenging because  both connections and nodes must be considered. For instance, simply contrasting graph adjacency information is not a meaningful similarity measure, as different edges can bear different importance in the graph structure. To calculate a distance that includes both nodes and edges, we propose a technique called Optimal Transport (OT), which transforms a (continuous) probability distribution into another with the lowest possible number of changes (effort). Therefore, given two graphs, the OT associated with their Wasserstein discrepancy (Barbe et al., 2021; Peyré et al., 2016) provides a correspondence between their nodes to establish graph matching (Alvarez-Melis \& Fusi, 2020; Barbe et al., 2020; Dong \& Sawin, 2020; Petric et al., 2019; Xu et al., 2019). Henceforth, when mentioning graph comparison, we will be referring to these techniques.

Graph comparison serves to avoid storing duplicated data at each time-step, that is, when WM is populated with the current state information (Algorithm 2). It also aids in retrieving comparable information that can be applied to the current state when performing matching between the current state and LTM concepts (Algorithm 3).

Once the memory system is populated with concepts, knowledge can be retrieved from LTM for use in the current state. Agent exploration provides the means to initialise LTM. Then, actions can be selected by processes of selective matching on the initialised memory using a trainable policy network. In this stage, the agent's focus shifts towards exploitation, that is, to achieve the goal by using the objects available in the environment or by building new ones. The next section will cover the latter scenario, describing the final Blending component.

Applying a concept from memory to a given state entails two steps: first, a match of the graph representing the state (e.g., \textit{$G_{t}$}) to the graph representing the long-term stored information (e.g., \textit{LTM}); second, once the match is successful, \textit{$G_{t}$} is supplemented with the nodes and edges (objects and affordances) present in the retrieved \textit{LTM subgraph} but lacking in \textit{$G_{t}$} in a process known as completion (Shanahan \& Mitchell, 2022) (Algorithm 4). Such completion will foster learning and bring to the agent's current state useful past experiences.

\begin{algorithm}
	\caption{Selective matching} 
	\begin{algorithmic}[1]

\State Let $G_{t}$,ltm, $s_{t}$, $Z$ be the input
\State $ltm-subgraphs \leftarrow graph-matching(G_{t}, ltm,Z)$ Where graph-matching is a function that matches a given graph against LTM and returns a list of sub-graphs from LTM with a Z\% similarity to $G_{t}$
\State $graph-union$ $\leftarrow$ union($ltm-subgraphs$) Where graph-union is a function that returns the union of a list of graphs
\State Return graph-union
	\end{algorithmic} 
\end{algorithm}

In itself, the state's information is a high-level abstraction of the graph formed by the objects and affordances at a lower level. This abstraction constitutes the node of a higher-level graph (state nodes). Edges between state nodes represent the temporal order of the frame sequence with an attached reward value.

Such a high-level graph formed by multiple state nodes populates the LTM unit, with each node being assigned a label (e.g.,1 or 0), depending on whether they have contributed to a successful episode (followed by a reward) or to a failed episode.

So far, we have shown how an agent could solve a problem using different existing concepts. The following component describes the process of creating a completely new concept by leveraging the existing concept space - which is considered (everyday) creativity.

\subsection{Blending component}
Creativity is triggered by situations where existing concepts are insufficient to solve the task at hand, leading an agent to a standstill. We are using the expression \textit{impasse situation} (Laird et al., 2012) to refer to the inability of an agent to solve a task for several episodes. A CPS approach should aid in overcoming such an impasse with a solution that satisfies problem constraints. Hence, we are working under the assumption that the impasse encountered when existing concepts are not suitable to solve the task could be surmounted by coming up with a novel, useful concept. In the AIGenC framework, new concepts are generated by blending existing concepts in the latent space into new representations. Concept blending is a term borrowed from Fauconnier \& Turner (1998) that denotes a process of combining meaningful features of two or more concepts into a new concept.

Two issues are to be addressed when creating a new concept. The first pertains to the selection of knowledge needed to generate meaningful information. The second is how to combine these concepts in a useful manner. A concept is helpful if it serves the agent to achieve its goal. To filter the concept space for information that can solve a problem, the agent needs to possess a high-level knowledge of the problem, the context and the task requirements; that is, to acquire a general problem template to process the information. General problem templates can be formed by grouping concepts into problem representations. This process requires summarising and assorting information at an abstraction layer above what the current hierarchy allows. Artificial agents lack this knowledge, and thus we propose a heuristic consisting of enlarging the range of matched concepts by lessening the similarity constraint. Hence, AIGenC's Blending component also performs a matching process, rendering the Reflective Reasoning and Blending units functionally complementary.

\begin{figure}[H]
   \centering
   \includegraphics[width=\textwidth]{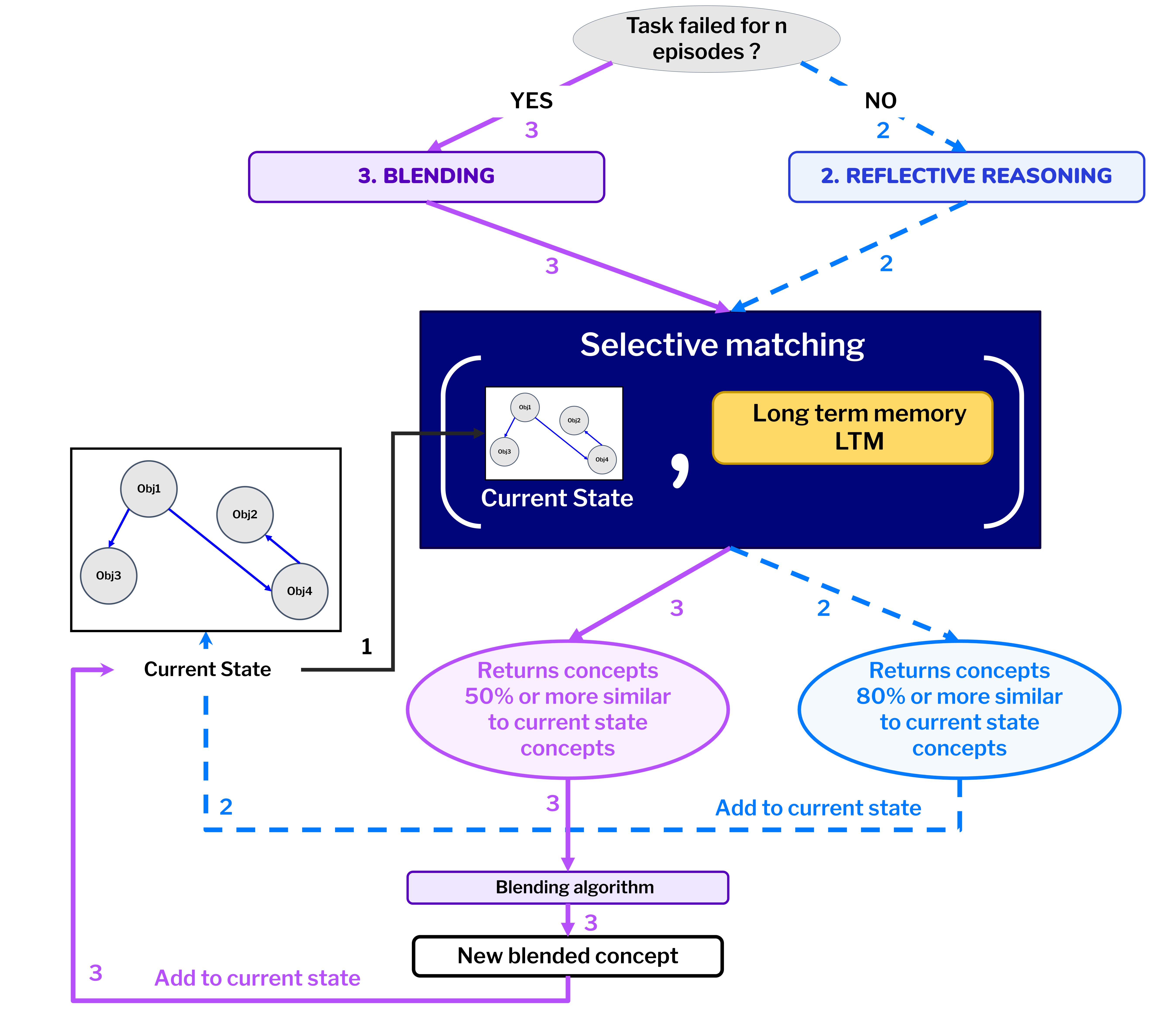}
   \caption{Reflective Reasoning and Blending complementary pipeline. Failure in solving a task on multiple episodes (grey top ellipse) activates Blending (3, left elements connected by solid pink lines); otherwise, the processing is carried out in Reflective Reasoning (2, right elements connected by dashed blue lines). A process of Selective Matching between the current state and Long Term Memory follows. Reflective reasoning selects concepts with a high similarity ratio and applies a union function to supplement the current state graphs. Blending selects concepts with a lower similarity rate. A Blending algorithm is run. The outcome of this algorithm is applied to the current state.}
   \label{fig:blending_rr}
\end{figure}

When the process of Reflective Reasoning, as described in the previous section, finds the concepts that lead to a satisfactory solution, the problem-solving task stops; otherwise, Blending is activated (Fig.\ref{fig:blending_rr}). Procedurally, both components begin operations by matching information from LTM to the current state. But, whereas in Reflective Reasoning, concepts are retrieved when there is a good match between nodes or edges of the current state and concepts stored in LTM, the Blending operation relaxes the similarity constraint to enlarge the pool of information (i.e., concepts), which permits operating on a larger variety of environmental data. 
Unlike in the Reflective Reasoning process, retrieved concepts are not used to supplement the current state with incremental information (union of existing LTM concepts) but rather with qualitative novel concepts obtained using a trainable network that combines information in a non-linear manner. Their combination contributes to expanding the concept space with new diverse concepts to be applied to the task.

To effectively blend latent representations of concepts using a trainable network, it is important to respect the structure of the latent space. This means that the dimensions of the latent vectors should be maintained, but the network should be able to combine their feature values by moving across their dimensions. To evaluate which features extracted by both the object discovery and the action selection sub-modules are most relevant and, therefore, needed in the combined new concept, a potential approach could be to quantify post hoc how much individual features contributed to each of the previous unsupervised models' performance. SHAP (Shapley Additive Explanations, Lundberg \& Lee, 2017), an interpretability method used to credit the predictive value of individual features, could offer a way to weigh their pertinence; the scores returned by the application of this method would thus be cardinal in selecting the new latent space features that bear most relevance to solving the task.

While generative models are commonly used to create new representations, they may not be suitable for producing new concepts. The classic Generative Adversarial Networks (GANs) (Goodfellow, 2016) formulation operates on a simple continuous input noise vector, imposing no restrictions on how the generator uses it. As a result, there is no guarantee that the generator would keep the dimensions of the initial vector that correspond to the semantic features of the data. A way to address this issue could be decomposing the input into two parts, an input vector and a latent code that will target the salient semantic features of the data distribution instead of employing a standard noise vector in a process akin to Chen et al.'s (2016). Thus, the input vector would be replaced by the concepts to be merged with a set of trainable parameters for their combination function. In addition, the latent codes could be initialised with an average of the SHAP values.

As the Blend function is not designed to reproduce the input represented in the latent space but to generate a new concept with it, the GAN loss function would need to be altered, for instance, using a clustering-specific loss that returns a low value when the new concept can be added to an existing cluster of concepts (Mukherjee et al., 2019).

Other proposals for latent vector manipulation use an autoencoder trained to learn a latent representation of objects, e.g., a chair (Bidgoli \& Veloso, 2018). This representation could be altered manually, and the modification decoded back into an image. However, a drawback of combining and altering the values of vectors in the latent space in this manner is that the outcome is not predictable. For instance, in Bigdoli and Veloso's model, altering the latent features that determine the dimensionality of the vector representing chairs is not guaranteed to generate acceptable chairs. By enforcing clustering similarity through the loss function, as suggested above, we could reduce the variability of the new concept, providing a mechanism for more meaningful concept building.

While we propose a similarity criterion (say, X\%) to select concepts to be blended,  there might be other suitable approaches. However, any Blending mechanism would only be possible in a system that incorporates functional components like those included in this model. It must be understood that the underlying representational structure of the first component is essential to obtain a solid input representation (i.e., latent codes representing the relevant independent features for both concept and action vectors). In addition, the tools described in Reflective Reasoning for selecting, adding and filtering concepts in the concept space are capital to minimise computational costs and recycle useful concepts. Filtering should not be confused with a mere search of the conceptual space. It implies a certain level of understanding of the high-level characteristics of the concept affordance that previously led to a comparable solution. This knowledge, possibly acquired after repeated similar experiences, requires the agent to hold an abstract representation of a generalisable solution template. 
 
This rationale, of course, implies that the process of generation of abstractions is incremental within the life of an agent, as is the case in animals. Based on previous learning, a human infant attempting to build a slingshot may think they need a V-shaped object and some sort of elastic material to propel projectiles. Knowledge of the required objects is acquired throughout the child's experiences by trial and error and probably some parental reinforcing guidance. Such learning experiences translate into a concept space that incrementally includes hierarchies of progressively more abstract affordances, leading to identifying the crucial components of a working propeller.
However, our agent is at the early stage of acquiring information, and hence it needs first to form the required lower-level hierarchy of concepts from those existing in the latent space. Hence, our agent uses a semi-random process of selection of relevant objects. Our hope is that the information acquired at this stage would set the necessary base for incremental abstract representations.

With Blending, the complete architecture has been presented. We have covered all the steps in building the first hierarchy of concepts, from collection and storage to using and creating new concepts. The architecture proposed serves as a blueprint for designing more robust systems equipped with transferable and adaptable knowledge capable of solving problems creatively.

\section{Discussion}
In this paper, we have introduced a deep reinforcement learning conceptual framework for creative problem solving as a mechanism for the transfer of relational information - a necessary first step in the hierarchical building of high-level abstractions. The model, alongside its three components, is designed to provide an agent with the capability to represent and create new concepts and transfer these learnt representations across tasks adaptively. From environmental features, the model constructs a hierarchical concept space that includes objects, affordances and representations of their relationships. The concept space is represented as vectors of features extracted using deep learning models. Matching, a part of the Reflective Reasoning component, is the operation used to retrieve past concepts relevant to the current problem. When the concepts retrieved fail to produce a satisfactory result, concepts can be recombined (blended) to form new concept representations and validated in the same reinforcement learning setup. In turn, this relational level of representation (i.e., graphs of concepts and affordances) is encoded together with rewards into abstract concept states represented as nodes in a higher-level graph in which edges convey time. The temporal layer of abstraction and the lower-level representation of affordances encode knowledge about the dynamics of the environment. Affordances are critical concepts for creative problem-solving. Having information about the affordances of an object in different contexts and in relation to other objects gives an agent the ability to form multiple different hypotheses (Pezzulo, 2008). Therefore, knowledge of affordances offers an agent multiple views of a concept (related to different possible goals), allowing the transfer of learning between tasks and goals.

Pretraining of basal representation features serves two objectives. It saves computational resources and recognises the influence that previous learning, development and evolution exert on natural agency, setting the necessary prior structures and knowledge to address complex tasks. As in biological agency, effective cognition does not manifest on tabula rasa assumptions (Marcus, 2018). We aim, however, to present a learning mechanism capable of reducing the use of handcrafted features in artificial agents, enabling them to creatively and adaptatively transform existing knowledge. Thus, we have designed a framework that shapes a concept space with minimal external interference, with a focus on adaptability and transferability of learning to improve AI generalisation.

We are not entering into the debate on how representations are formed, nor do we aim to assess any of the claims regarding concept formation in humans -which remains an open question that sparks lively discussion (Piantadosi, 2021). Despite taking inspiration from cognitive theories, our interest lies exclusively in describing the necessary concept space and interplaying structures that could aid an artificial agent in transferring knowledge across contexts and tasks. To that objective, we have looked into several cognitive approaches. We have reused various ideas, from the connectionist approach to low-level concept formation to the hierarchical nature of the concept space proposed by theories of cognition.
In our work, we steer away from assigning anthropomorphic value to the different components and the different outputs each of them produces (Watson, 2019), but we acknowledge that such analogies help us frame learning strategies and have influenced our creative problem-solving proposal.

Whether our model can, in turn, shed some light to modelling in cognition is outside the scope of this proposal, but we hope that by offering cogent computational tools to simulate different processes, our framework can help refine the process of building accurate and cognitive plausible models.

Divergent currents of opinion regarding the prospect of AI to achieve human-like intelligence swarm academic fora, and scientific publications reflect the debate –from those that caution on conceptual missuses arising from interdisciplinary approaches (Shevlin \& Halina, 2019) and papers that epitomise trends that defend the implausibility of Artificial General Intelligence (AGI) (Fjelland, 2020; Korteling et al., 2021) to special issues dedicated to cognitive-inspired approaches to computing (Zhu et al., 2020). At the eye of the storm, the lack of AI generalisation has been highlighted as one of, if not the most crucial problem of AI (Crosby et al., 2019; Zhang et al., 2021). Humans and animals are capable to generalise across settings by detecting the features that they hold in common. Similarity, at different levels of abstraction, drives the transfer of knowledge. 

We concur with cognitive-inspired theories that posit that the lack of generalisation in AI agents is partially due to their inability to discover, combine and generate new concepts that can be used across domains to solve different, yet similar, tasks (Shanahan et al., 2020). After all, human experiences are constant, they are prolonged beyond childhood and the amount, quality and diversity of information received is probably more substantial than that fed into artificial agents. Moreover, unlike machines, animals and humans must learn to perceptually discern the flood of information, setting apart distinctive and shared features in a process driven by comparison mechanisms (Gibson, E. J., 1992; Mondragón \& Murphy, 2010). It is conceivable that this operation underpins learning so-called common sense knowledge. 

It is likely that the larger the repertoire of concepts and spatial and temporal information that an artificial agent has access to in a concept space, the more commonalities it can learn, commonalities that, if encoded at different levels of a hierarchy, can serve as a world frame and offer some level of "common sense" to the agent. In that vein, our proposal unfolds from contextualised functional representations. The AI system we propose will learn an adaptive concept space by interacting with the environment by trial and error to detect and differentiate commonalities, contextual information and unique functional features, thus enabling it to transfer knowledge more efficiently.

Should an implementation of our computational framework prove successful in generalising across domains and tasks, it could be argued that the main pending barrier in achieving AGI might begin to fracture. In such case, the reluctance of sceptics to acknowledge \textit{proper} intelligence in artificial systems would be principled in a purely anthropocentric stance.

\newpage
\section*{References} 
\begin{itemize}
\item[] Alvarez-Melis, D., \& Fusi, N.(2020). Geometric dataset distances via optimal transport. \textit{Advances in Neural Information Processing Systems, 33}, 21428-21439.
\item[] Badia, A.P., Piot, B., Kapturowski, S., Sprechmann, P., Vitvitskyi, A., Guo, Z.D. \& Blundell, C.. (2020). Agent57: Outperforming the Atari Human Benchmark. \textit{Proceedings of the 37th International Conference on Machine Learning}, in \textit{Proceedings of Machine Learning Research 119} (pp.507-517). Available from https://proceedings.mlr.press/v119/badia20a.html.
\item[] Baker, B., Kanitscheider, I., Markov, T., Wu, Y., Powell, G., McGrew, B., \& Mordatch, I.(2019). Emergent tool use from multi-agent autocurricula. \textit{arXiv preprint arXiv:1909.07528}.
\item[] Barbe, A., Sebban, M., Gonçalves, P., Borgnat, P., \& Gribonval, R.(2020). Graph diffusion Wasserstein distances. \textit{Proceedings of ECML PKDD 2020-European Conference on Machine Learning and Principles and Practice of Knowledge Discovery in Databases, Sep 2020}, (pp. 1-16).
\item[] Bayer, M., Kaufhold, M. A., Buchhold, B., Keller, M., Dallmeyer, J., \& Reuter, C. (2022). Data augmentation in natural language processing: a novel text generation approach for long and short text classifiers. \textit{International Journal of Machine Learning and Cybernetics, 1-16}.
\item[] Bellemare, M. G., Naddaf, Y., Veness, J., \& Bowling, M. (2013). The arcade learning environment: An evaluation platform for general agents. \textit{Journal of Artificial Intelligence Research, 47}, 253-279.
\item[] Bengio, Y., Lecun, Y., \& Hinton, G.(2021). Deep learning for AI. \textit{Communications of the ACM, 64(7)}, 58-65.
\item[] Beyret, B., Hernández-Orallo, J., Cheke, L., Halina, M., Shanahan, M., \& Crosby, M. (2019). The animal-ai environment: Training and testing animal-like artificial cognition. arXiv preprint arXiv:1909.07483.
\item[] Bidgoli, A., \& Veloso, A., (2018). DeepCloud. The Application of a Data-Driven, Generative Model in Design. In P. Anzalone, M. Del Signore, \& A.J. Wit (Eds.),\textit{ Recalibration: On Imprecision and Infidelity.  Proceedings of the 38th ACADIA Conference} (pp. 176–85). Mexico City: Universidad Iberoamericana. Available from http://papers.cumincad.org/cgi-bin/works/paper/acadia18\_176.
\item[] Chemero, A., Klein, C., \& Cordeiro, W. (2003). Events as changes in the layout of affordances. \textit{Ecological Psychology, 15(1)}, 19-28.
\item[] Chen, X., Duan, Y., Houthooft, R., Schulman, J., Sutskever, I., \& Abbeel, P. (2016). Infogan: Interpretable representation learning by information maximizing generative adversarial nets. \textit{Advances in neural information processing systems, 29}.
\item[] Chevalier-Boisvert, M., Willems, L., \& Pal, S. (2018). Minimalistic gridworld environment for openai gym.
\item[] Chollet, F.(2019). On the measure of intelligence. \\\textit{arXiv preprint arXiv:1911.01547}.
\item[] Colin, T.R., Belpaeme, T., Cangelosi, A., \& Hemion, N.(2016). Hierarchical reinforcement learning as creative problem solving. \textit{Robotics and Autonomous Systems, 86}, 196-206.
\item[] Coraci, D.(2022). A Unified Model of Ad Hoc Concepts in Conceptual Spaces. \textit{Minds and Machines, 1-21}.
\item[] Crosby, M., Beyret, B., \& Halina, M.(2019). The animal-ai olympics. \textit{Nature Machine Intelligence, 1(5)}, 257-257.
\item[] Danks, D., \& Plis, S. (2019). Amalgamating evidence of dynamics. \textit{Synthese, 196(8)}, 3213-3230.
\item[] Dasgupta, I., Guo, D., Gershman, S. J., \& Goodman, N. D. (2020). Analyzing machine‐learned representations: A natural language case study. \textit{Cognitive Science, 44(12)}.
\item[] Dong, Y., \& Sawin, W.(2020). COPT: Coordinated optimal transport on graphs. \textit{Advances in Neural Information Processing Systems, 33}, 19327-19338.
\item[] Doumas, L.A., Hummel, J.E., \& Sandhofer, C.M.(2008). A theory of the discovery and predication of relational concepts. \textit{Psychological Review, 115(1)}, 1.
\item[] Doumas, L.A.,Puebla, G.,Martin, A.E., \& Hummel, J.E.(2022). A theory of relation learning and cross-domain generalisation. \textit{Psychological Review}. Advance online publication.
\item[] Edwards, H., \& Storkey, A.(2016). Towards a neural statistician.  \textit{arXiv preprint arXiv:1606.02185}.
\item[] Fauconnier, G. \& Turner, M.(1998). Conceptual integration networks. \textit{Cognitive Science, 22}, 133-187.
\item[] Fjelland, R.(2020). Why general artificial intelligence will not be realised. \textit{Humanities and Social Sciences Communications, 7(1)}, 1-9.
\item[] Fletcher, L., \& Carruthers, P.(2012). Metacognition and reasoning. \textit{Philosophical Transactions of the Royal Society B: Biological Sciences, 367(1594)}, 1366-1378.
\item[] Floridi, L.,\& Chiriatti, M.(2020). GPT-3: Its nature, scope, limits, and consequences. \textit{Minds and Machines, 30(4)}, 681-694.
\item[] Frith, E., Elbich, D.B., Christensen, A.P., Rosenberg, M.D., Chen, Q., Kane, M.J., Silvia, P.J., Seli, P., Beaty, R.E.(2021). Intelligence and creativity share a common cognitive and neural basis.  \textit{Journal of Experimental Psychology: General, 150(4)}, 609.
\item[] Gärdenfors, P.(2004). \textit{Conceptual spaces: The geometry of thought}. MIT press.
\item[] Geirhos, R., Jacobsen, J.K., Michaelis, C., Zemel, R., Brendel, W., Bethge, M., Wichmann, F.A.(2020). Shortcut learning in deep neural networks. \textit{Nature Machine Intelligence,2(11)}, 665-673.
\item[] Gibson, E. J. (1992). How to think about perceptual learning: Twenty-five years later. In H. L. Pick, Jr., P. W. van den Broek, \& D. C. Knill (Eds.), Cognition: Conceptual and methodological issues (pp. 215–237). American Psychological Association. https://doi.org/10.1037/10564-009
\item[] Gibson, J.J.(1977). \textit{The theory of affordances}. Hilldale, USA.
\item[] Gizzi, E., Nair, L., Sinapov, J., \& Chernova, S.(2020). From Computational Creativity to Creative Problem Solving Agents. \textit{Proceedings of the 11th International Conference on Computational Creativity, (pp. 370-373)}
\item[] Goodfellow, I. (2016). Nips 2016 tutorial: Generative adversarial networks. arXiv preprint arXiv:1701.00160.
\item[] Graves, A., Wayne, G., \& Danihelka, I.(2014). Neural turing machines. \textit{arXiv preprint arXiv:1410.5401}.
\item[] Guilford, J.P.(1967). \textit{The nature of human intelligence}.  McGraw-Hill.
\item[] Haskell, R.E.(2000). \textit{Transfer of learning: Cognition and instruction (p.30)}. Elsevier.
\item[] Hassanin, M., Khan, S., \& Tahtali, M. (2021). Visual affordance and function understanding: A survey. \textit{ACM Computing Surveys (CSUR), 54(3)}, 1-35.
\item[] Hayman, G., \& Huebner, B.(2019). Temporal updating, behavioral learning, and the phenomenology of time-consciousness. \textit{Behavioral and Brain Sciences, 42}.
\item[] Jacobs, M.K., \& Dominowski, R.L.(1981). Learning to solve insight problems. \textit{Bulletin of the Psychonomic Society, 17(4)}, 171-174.
\item[] Jaderberg, M., Mnih, V., Czarnecki, W.M., Schaul, T., Leibo, J. Z., Silver, D., \& Kavukcuoglu, K.(2016). Reinforcement learning with unsupervised auxiliary tasks. \textit{arXiv preprint arXiv:1611.05397}.
\item[] Jain, A., Szot, A. \& Lim J.J.(2020). Generalisation to New Actions in Reinforcement Learning. \textit{Proceedings of the 37 the International Conference on Machine Learning, Online}, PMLR 119.
\item[] James, W.(1890). \textit{The principles of psychology (p. 251)}. Henry Holt and co.
\item[] Johnson, J., Hariharan, B., Van Der Maaten, L., Fei-Fei, L., Lawrence Zitnick, C., \& Girshick, R.(2017). Clevr: A diagnostic dataset for compositional language and elementary visual reasoning. \textit{Proceedings of the IEEE conference on computer vision and pattern recognition (pp. 2901-2910)}.
\item[] Kingma, D.P., \& Welling, M.(2013). Auto-encoding variational bayes. \textit{arXiv preprint arXiv:1312.6114}.
\item[] Korteling, J., van de Boer-Visschedijk, G., Blankendaal, R., Boonekamp, R., \& Eikelboom, A.(2021).Human-versus artificial intelligence. \textit{Frontiers in artificial intelligence, 4}.
\item[] Laird, J. E., Derbinsky, N., \& Tinkerhess, M.(2012). Online determination of value-function structure and action-value estimates for reinforcement learning in a cognitive architecture. \textit{Advances in Cognitive Systems, 2}, 221-238.
\item[] LeCun, Y., Bottou, L., Bengio, Y., \& Haffner, P.(1998). Gradient-based learning applied to document recognition. \textit{Proceedings of the IEEE, 86(11)}, (pp. 2278-2324).
\item[] Locatello, F., Weissenborn, D., Unterthiner, T., Mahendran, A., Heigold, G., Uszkoreit, J., Dosovitskiy, A., \& Kipf, T.(2020). Object-centric learning with slot attention. \textit{Advances in Neural Information Processing Systems, 33}, 11525-11538.
\item[] Loynd, R., Fernandez, R., Celikyilmaz, A., Swaminathan, A., \& Hausknecht, M.(2020). Working memory graphs. \textit{Proceedings of the International Conference on Machine Learning (pp. 6404-6414)}. PMLR.
\item[] Lundberg, S. M., \& Lee, S. I. (2017). A unified approach to interpreting model predictions. \textit{Advances in neural information processing systems, 30}.
\item[] Lyre, H.(2020). The State Space of Artificial Intelligence. \textit{Minds and Machines,30(3)}, 325-347.
\item[] Marcus, G. (2018). Innateness, alphazero, and artificial intelligence. arXiv preprint arXiv:1801.05667.
\item[] Masci, J., Meier, U., Cireşan, D., \& Schmidhuber, J.(2011, June). Stacked convolutional auto-encoders for hierarchical feature extraction. \textit{Proceedings of the International conference on artificial neural networks (pp. 52-59)}. Springer.
\item[] Mednick, S.(1962). The associative basis of the creative process. \textit{Psychological Review, 69(3)}, 220–232.
\item[] Mezghani, L., Sukhbaatar, S., Lavril, T., Maksymets, O., Batra, D., Bojanowski, P., \& Alahari, K.(2021). Memory-augmented reinforcement learning for image-goal navigation. \textit{arXiv preprint arXiv:2101.05181}.
\item[] Mitchell, M.(2021). Abstraction and analogy‐making in artificial intelligence. \textit{Annals of the New York Academy of Sciences,1505(1)},79-101.
\item[] Mitra, P.P.(2021). Fitting elephants in modern machine learning by statistically consistent interpolation. \textit{Nature Machine Intelligence, 3(5)}, 378-386.
\item[] Momennejad, I.(2020). Learning structures: predictive representations, replay, and generalisation. \textit{Current Opinion in Behavioral Sciences, 32}, 155-166.
\item[] Mondragón, E. \& Murphy, R. A. (2010). Perceptual learning in appetitive conditioning: Analysis of the Effectiveness of the Common Element. Behavioural Processes, 83, 247-256. doi: 10.1016/j.beproc.2009.12.007.
\item[] Mondragón, E., Alonso, E. \& Kokkola, N.(2017). Associative learning should go deep. \textit{Trends in Cognitive Sciences 21(11)}, 822-825.
\item[] Mukherjee, S., Asnani, H., Lin, E., \& Kannan, S. (2019). Clustergan: Latent space clustering in generative adversarial networks. \textit{In Proceedings of the AAAI conference on artificial intelligence, 33}, 4610-4617.
\item[] Murphy, R.A., Mondragón, E., \& Murphy, V.A.(2008). Rule learning by rats. \textit{Science, 319(5871)}, 1849-1851.
\item[] Nickerson, R.S.(1987). \textit{Why teach thinking?}. W H Freeman/Times Books/ Henry Holt \& Co.
\item[] Oltețeanu, A.M.(2020). \textit{Cognition and the Creative Machine: Cognitive AI for Creative Problem Solving (p.92)}. Springer Nature.
\item[] Petric Maretic, H., El Gheche, M., Chierchia, G., \& Frossard, P.(2019). GOT: an optimal transport framework for graph comparison. \textit{Advances in Neural Information Processing Systems, 32}.
\item[] Pezzulo, G.(2008). Coordinating with the future: the anticipatory nature of representation. \textit{Minds and Machines, 18(2)},179-225.
\item[] Piantadosi, S.T.(2021). The computational origin of representation. \textit{Minds and machines, 31(1)}, 1-58.
\item[] Radulescu, A., Shin, Y.S., \& Niv, Y.(2021). Human representation learning. \textit{Annual Review of Neuroscience, 44}, 253-273.
\item[] Rae, J.W., Borgeaud, S., Cai, T., Millican, K., Hoffmann, J., Song, F., ... \& Irving, G.(2021). Scaling language models: Methods, analysis \& insights from training gopher. \textit{arXiv preprint arXiv:2112.11446}.
\item[] Raileanu, R., Goldstein, M., Yarats, D., Kostrikov, I., \& Fergus, R.(2021). Automatic Data Augmentation for Generalization in Reinforcement Learning. \textit{Advances in Neural Information Processing Systems}, 34.
\item[] Raji, I. D., Bender, E. M., Paullada, A., Denton, E., \& Hanna, A. (2021). AI and the everything in the whole wide world benchmark. arXiv preprint arXiv:2111.15366.
\item[] Ramesh, A., Pavlov, M., Goh, G., Gray, S., Voss, C., Radford, A., Chen, M., Sutskever, I.(2021). Zero-shot text-to-image generation. \textit{Proceedings of the International Conference on Machine Learning (pp. 8821-8831)}.PMLR.
\item[] Reed, S., Zolna, K., Parisotto, E., Colmenarejo, S. G., Novikov, A., Barth-Maron, G., ... \& de Freitas, N.(2022). A Generalist Agent. \textit{arXiv preprint arXiv:2205.06175}.
\item[] Rezende, D.J., Mohamed, S., \& Wierstra, D.(2014). Stochastic backpropagation and approximate inference in deep generative models. \textit{Proceedings of the International conference on machine learning (pp. 1278-1286)}. PMLR.
\item[] Rumelhart, D.E., Hinton, G.E. and Williams, R.J. (1986). \textit{Learning Internal Representations by Error Propagation. In D. E. Rumelhart, J. L., McClelland and the PDP Research Group (Eds.), Parallel Distributed Processing: Explorations in the Microstructure of Cognition, Vol. 1: Foundations, (318-362)}. MIT Press.
\item[] Russell, S., \& Norvig, P.(2002). \textit{Artificial intelligence: a modern approach (Chapter.8)}. Pearson Education Limited.
\item[] Şahin, E., Cakmak, M., Doğar, M. R., Uğur, E., \& Üçoluk, G.(2007). To afford or not to afford: A new formalization of affordances toward affordance-based robot control. \textit{Adaptive Behavior, 15(4)}, 447-472.
\item[] Searle, J.R.(1980). Minds, brains, and programs. \textit{Behavioral and brain sciences, 3(3)}, 417-424.
\item[] Shanahan, M., Crosby, M., Beyret, B. \& Cheke, L.(2020). Artificial Intelligence and the Common Sense of Animals. \textit{Trends in Cognitive Sciences,24(11)}, 862-872.
\item[] Shanahan, M.,\& Mitchell, M.(2022). Abstraction for Deep Reinforcement Learning. \textit{arXiv preprint arXiv:2202.05839}.
\item[] Shevlin, H., \& Halina, M.(2019). Apply rich psychological terms in AI with care. \textit{Nature Machine Intelligence, 1(4)}, 165-167.
\item[] Sloutsky, V. M.(2010). From perceptual categories to concepts: What develops?. \textit{Cognitive science, 34(7)}, 1244-1286.
\item[] Sridhar, M., Cohn, A. G., \& Hogg, D. C. (2008). Learning functional object categories from a relational spatio-temporal representation. \textit{Proceedings of ECAI 2008: 18th European Conference on Artificial Intelligence (Frontiers in Artificial Intelligence and Applications) (pp. 606-610)}.
\item[] Sridhar, M., Cohn, A.G., \& Hogg, D.C. (2010). Unsupervised learning of event classes from video. \textit{Proceedings of the Twenty-Fourth AAAI Conference on Artificial Intelligence} (pp. 1631-1638.
\item[] Stooke, A., Lee, K., Abbeel, P., \& Laskin, M.(2021). Decoupling representation learning from reinforcement learning. \textit{Proceedings of the International Conference on Machine Learning (pp. 9870-9879)}. PMLR.
\item[] Sutton, R.S., \& Barto, A.G.(2018). \textit{Reinforcement learning: An introduction}. MIT press.
\item[] Villani, C.(2009). \textit{Optimal transport: old and new (Vol. 338)}. Berlin: springer.
\item[] Vinyals, O., Babuschkin, I., Czarnecki, W.M., Mathieu, M., Dudzik, A., Chung, J., ... \& Silver, D.(2019). Grandmaster level in StarCraft II using multi-agent reinforcement learning. \textit{Nature, 575(7782)}, 350-354.
\item[] Wallas, G.(1926). \textit{The art of thought}. Harcourt, Brace.
\item[] Watson, D.(2019). The rhetoric and reality of anthropomorphism in artificial intelligence. \textit{Minds and Machines, 29(3)}, 417-440.
\item[] Wiggins, G.A.(2006). A preliminary framework for description, analysis and comparison of creative systems. \textit{Knowledge-Based Systems, 19(7)}, 449-458.
\item[] Wu, C.M., Schulz, E., Speekenbrink, M., Nelson, J.D., \& Meder, B.(2018). Generalization guides human exploration in vast decision spaces. \textit{Nature human behaviour, 2(12)}, 915-924.
\item[] Xie, S., Ma, X., Yu,P., Zhu,Y., Wu, Y.N., \& Zhu, S.C. (2021). Halma: Humanlike abstraction learning meets affordance in rapid problem solving. \textit{arXiv preprint arXiv:2102.11344}.
\item[] Xu, H., Luo, D., \& Carin, L.(2019). Scalable Gromov-Wasserstein learning for graph partitioning and matching. \textit{Advances in neural information processing systems, 32 (NeurIPS 2019)}.
\item[] Yang, Y., Cao, J., Wen, Y., \& Zhang, P. (2021). Table to text generation with accurate content copying. \textit{Scientific reports, 11(1), 1-12}.
\item[] Yarats, D., Fergus, R., Lazaric, A., \& Pinto, L.(2021). Reinforcement learning with prototypical representations. \textit{Proceedings of the International Conference on Machine Learning (pp. 11920-11931}). PMLR.
\item[] Zhang, C., Bengio, S., Hardt, M., Recht, B., \& Vinyals, O.(2021). Understanding deep learning (still) requires rethinking generalisation. \textit{Communications of the ACM, 64(3)}, 107-115.
\item[] Zhang, W., GX-Chen, A., Sobal, V., LeCun, Y., \& Carion, N. (2022). Light-weight probing of unsupervised representations for Reinforcement Learning. arXiv preprint arXiv:2208.12345.
\item[] Zhu, R., Liu, L., Ma, M., \& Li, H.(2020). Cognitive-inspired computing: Advances and novel applications. \textit{Future Generation Computer Systems, 109}, 706-709.
\end{itemize}

\end{document}